\documentclass[runningheads]{llncs}
\usepackage{graphicx}
\usepackage{multirow}
\usepackage{comment} 
\usepackage{color}

\begin{document}
\title{Point Cloud Novelty Detection Based on Latent Representations of a General Feature Extractor
}

\author{Shizuka Akahori\inst{1} \and 
Satoshi Iizuka\inst{2,3}\orcidID{0000-0001-9136-8297} \and
Ken Mawatari\inst{4,5}\orcidID{0000-0003-4985-0201} \and
Kazuhiro Fukui\inst{2,3}\orcidID{0000-0002-4201-1096} }

\authorrunning{S. Akahori et al.} 

\institute{
Graduate School of Science and Technology, University of Tsukuba, 1-1-1 Tennodai, Tsukuba, Ibaraki 305-8571, Japan\\
\email{akahori.shizuka@image.iit.tsukuba.ac.jp} \and 
Institute of Systems and Information Engineering, University of Tsukuba, 1-1-1 Tennodai, Tsukuba, Ibaraki 305-8571, Japan \and
Center for Artificial Intelligence Research, University of Tsukuba, 1-1-1 Tennodai, Tsukuba, Ibaraki 305-8571, Japan\\
\email{\{iizuka,kfukui\}@cs.tsukuba.ac.jp} \and 
Division of Physics, Faculty of Pure and Applied Sciences, University of Tsukuba, 1-1-1 Tennodai, Tsukuba, Ibaraki 305-8571, Japan \and 
Tomonaga Center for the History of the Universe (TCHoU), Faculty of Pure and Applied Sciences, University of Tsukuba, 1-1-1 Tennodai, Tsukuba, Ibaraki 305-8571, Japan\\
\email{mawatari.ken.ka@u.tsukuba.ac.jp} \\
}

\maketitle              
\begin{abstract}
We propose an effective unsupervised 3D point cloud novelty detection approach, leveraging a general point cloud feature extractor and a one-class classifier. The general feature extractor consists of a graph-based autoencoder and is trained once on a point cloud dataset such as a mathematically generated fractal 3D point cloud dataset that is independent of normal/abnormal categories. The input point clouds are first converted into latent vectors by the general feature extractor, and then one-class classification is performed on the latent vectors. Compared to existing methods measuring the reconstruction error in 3D coordinate space, our approach utilizes latent representations where the shape information is condensed, which allows more direct and effective novelty detection. We confirm that our general feature extractor can extract shape features of unseen categories, eliminating the need for autoencoder re-training and reducing the computational burden. We validate the performance of our method through experiments on several subsets of the ShapeNet dataset and demonstrate that our latent-based approach outperforms the existing methods.

\end{abstract}

\section{Introduction}
Novelty detection, also known as out-of-distribution detection or unsupervised anomaly detection, is a critical classification task to identify anomalous patterns that differ from trained data distribution~\cite{novelty,ND}. Although deep neural networks have improved recognition tasks in recent years, their performance is susceptible to novel data not included in the training set. In particular, given data that does not belong to any training classes, the trained classifier network will force the given data to be classified into one of the predefined classes, causing serious classification errors. In this situation, novelty detection can be applied as a preprocessing step to eliminate such anomalous data, thereby enhancing the reliability of the classification results. While various methods have been proposed for novelty detection in 2D images, limited research has focused on novelty detection in 3D point clouds, which is essential to improve 3D classification performance in industrial systems.

Recently, Masuda et al.~\cite{keio} proposed a novelty detection method using a variational autoencoder~\cite{VAE} by measuring the reconstruction loss of 3D point clouds as the anomaly score. While it is the first method to tackle unsupervised anomaly detection in 3D point clouds, the detection performance depends on the quality of the decoder network, and the anomaly score measured in 3D coordinate space tends to be sensitive to detailed object geometry, sampling, and data noise as mentioned in~\cite{dpd}. Moreover, different distribution or classes of normal data requires the encoder network re-training process, which results in high computational costs.

To address these issues, we propose an unsupervised 3D point novelty detection approach comprising a general point cloud feature extractor and a one-class classifier. As the general feature extractor, we first train an autoencoder (AE)~\cite{AE} on the dataset of which the data is agnostic to the normal and anomaly class categories. Next, by the feature extractor, the training data of only normal classes are converted into latent vectors to extract condensed shape information. Then, one-class classification such as One-Class Support Vector Machine (OC-SVM)~\cite{OCSVM} and Kernel PCA-based Novelty Detection (KPCA-ND)~\cite{KPCA-ND} is trained on the latent vectors of the training data. For testing, an input point cloud is also transformed into a latent vector and measured its abnormality by the one-class classification. We confirm that our general feature extractor has the ability to extract the shape information of even unseen classes, which is beneficial for effective novelty detection and lower computational expenses. We validate our approach on three types of point cloud datasets and compare it with the state-of-the-art method. The results demonstrate that our latent-based OC-SVM and KPCA-ND exhibit significant improvements in point cloud novelty detection. In addition, we compared the performance of general feature extractors trained on different datasets while visualizing latent variables.

\section{Related work}
There have been various approaches for novelty detection of 2D images~\cite{NDImage0,NDImage1,NDImage2,NDImage3,NDImage4,NDImage5}. Although these methods show high performance, they focus only on image data and do not handle 3D point cloud data. Unlike 2D image data, where each pixel has color information, 3D point cloud data has position information. In this work, we propose an effective framework for novelty detection of point cloud data.

\subsection{Novelty Detection of 3D Point Clouds} 
Recently, Masuda et al.~\cite{keio} proposed a reconstruction-based method measuring the anomaly score between the input and the reconstructed point cloud using a Variational AutoEncoder (VAE)~\cite{VAE}. Qin et al.~\cite{teacherstudent} introduced a teacher-student network that minimizes a multi-scale loss between the feature vectors generated by the teacher and student networks. While the student network can be trained on a few training data, the performance is sensitive to the selection of the training samples especially in cases where the dataset contains many sparse point clouds like ShapeNet~\cite{ShapeNet}.

\textbf{Feature learning of 3D point clouds}.
In order to extract 3D shape features from point clouds, deep feature learning approaches are proposed, including supervised-based~\cite{PointNet,PointNet2,so-net}, self-supervised-based~\cite {pointcontrast,SSL-1,SPU-Net}, and unsupervised-based~\cite{foldingnet,MAP-VAE}. 
While the recent image anomaly detection tasks often utilize the general feature extractor such as ResNet~\cite{resnet} trained on ImageNet~\cite{imagenet}, it is less common to use such trained feature extractors in point cloud anomaly detection tasks. One factor behind this is the lack of publically available large-scale point cloud datasets. Recently, a novel method for generating training data with natural 3D structures using fractal geometries has been proposed~\cite{Fractal}.

\subsection{One-class Classification}
Various one-class classification methods~\cite{nearest,k-nn,SVDD,DeepSVDD,OCSVM,PCA,KPCA-ND} have been developed for image analysis, sensors, and signal monitoring. 
OC-SVM is the method to maximize the margin of the discriminative hyperplane from the origin, where the data points located outside the hyperplane are considered anomalies or novelties. The use of kernel functions can improve discrimination by projecting data into a higher dimensional space.  Deep Support Vector Data Description (DeepSVDD)~\cite{DeepSVDD} is the deep learning-based extension of the SVDD~\cite{SVDD} technique that identifies a hypersphere enclosing the training data with a minimum possible radius. It trains a neural network to map the input data close to the center of the hypersphere. 
Generalised One-class Discriminative Subspaces (GODS)~\cite{GODS} learns a set of discriminative hyperplanes to bound normal data, where each hyperplane consists of an orthogonal subspace. GODS can combine the linearity of OC-SVM with the nonlinear boundary properties of SVDD. 
KPCA-ND~\cite{KPCA-ND} is a nonlinear classification method that utilizes Kernel PCA~\cite{kernelpca}. Training data is mapped into an infinite-dimensional feature space using the kernel function such as Gaussian kernel, and PCA is performed on the mapped training data to compute eigenvectors. The cosine similarity between the eigenvectors and the test data that is mapped into the same feature space is measured.

\begin{figure}
\centering
\includegraphics[width=0.9\textwidth]{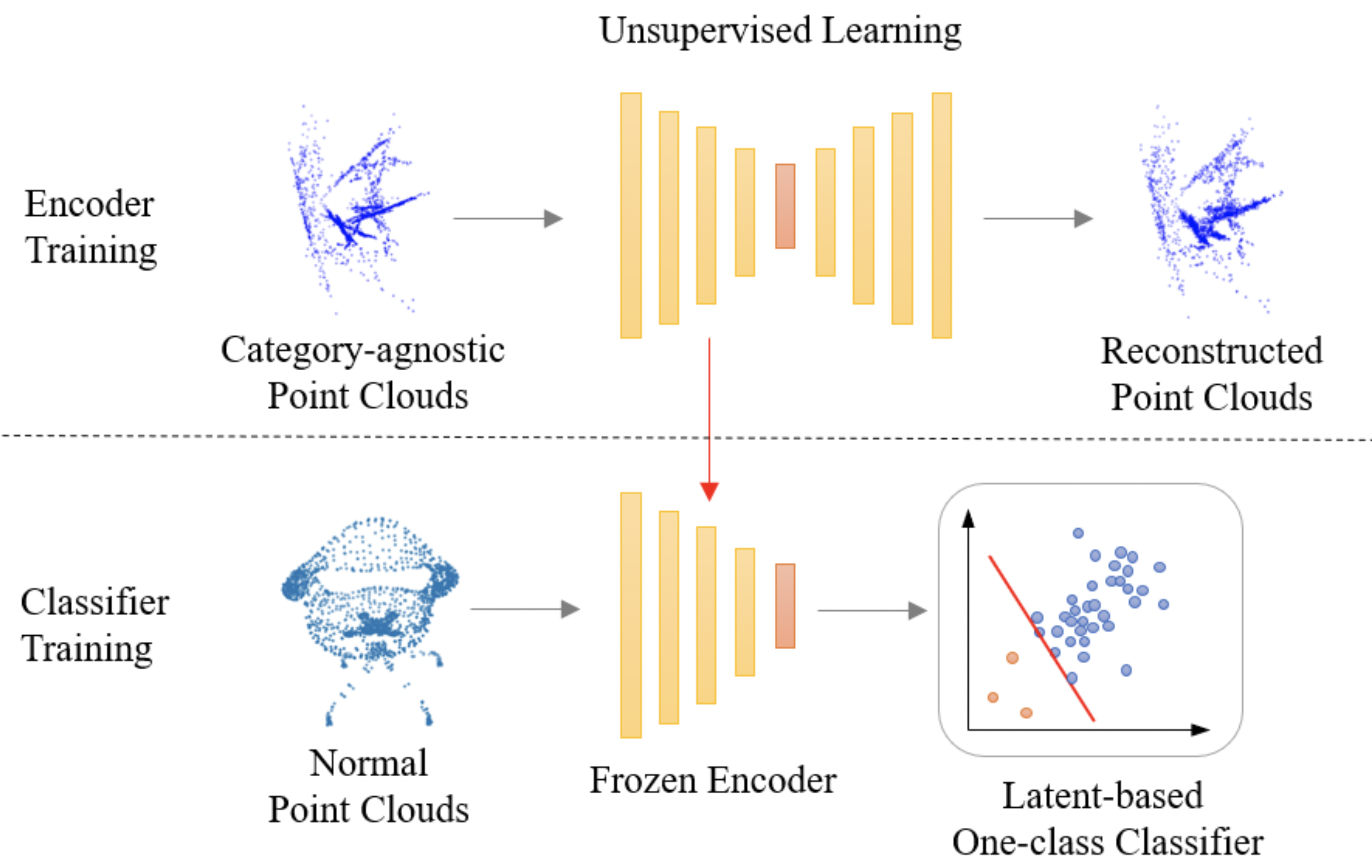}
\caption{Our latent-based one-class classification approach. 
 Composed of a general point cloud feature extractor and a one-class classifier. As the general feature extractor, an autoencoder is trained on the dataset that is independent of detection categories. Input point clouds are converted into latent vectors by the trained frozen autoencoder. A one-class classifier trained on the latent vectors of normal classes classifies input as normal or anomaly.} 
\label{fig:overview}
\vspace{-0.2cm}  
\end{figure}

\section{Proposed Framework}
Our framework comprises two main components: a general point cloud feature extractor and a one-class classifier. First, the autoencoder is trained on a dataset that is independent of normal and abnormal categories. This training process is performed only once, and the trained autoencoder functions as a general feature extractor that can be used independently of specific test classes and distributions. Second, both training and test data are fed into the trained autoencoder, and the latent vectors are extracted at the bottleneck of the autoencoder. Third, the one-class classifier is trained on the extracted latent vectors of training data consisting only of normal data, then is used to classify whether the test data is normal or anomaly via a trained one-class classifier. In contrast to existing reconstruction-based approaches that measure anomalies in 3D coordinate space, our method operates in latent space where geometric features are compressed, allowing for more effective and robust novelty detection. In addition, once the autoencoder is trained, the weights are fixed, and the frozen autoencoder can be used as the general feature extractor, which eliminates the need for the encoder re-training and is effective in reducing computational costs. The training of both the encoder and classifier is performed in unsupervised learning.

\subsection{Point Cloud Feature Extractor}
We use a graph-based autoencoder to extract a discriminative feature representation from the input 3D point cloud data. The autoencoder is based on the FoldingNet architecture~\cite{foldingnet} similar to that of Masuda et al.~\cite{keio}, but it differs in using a standard autoencoder instead of a variational autoencoder. The standard autoencoder performs better in the area under the curve (AUC) metric than the variational version since the latent vectors are more dispersed as shown in Figure~\ref{fig:m10}(a), Figure~\ref{fig:VAE_PointNet}(a).

The point cloud autoencoder is composed of an encoder $E$ with learnable parameters $\theta_e$ and decoder $D_1$, $D_2$ with learnable parameters $\theta_{d1}$, $\theta_{d2}$ which is trained to reconstruct the input 3D point cloud $S$. The reconstruction loss $L$ between the input 3D point cloud $S$ and the reconstructed 3D point cloud $\hat{S}$ is measured by Chamfer Distance $d_{CD}$~\cite{chamfer}. This is represented as follows:



\begin{equation}
    d_{CD}(P, Q) = \max\biggr\{\frac{1}{|P|}\sum_{p\in P}\min_{q\in Q} \|p-q\|_2 + \frac{1}{|Q|}\sum_{q\in Q}\min_{p\in P} \|q-p\|_2\biggr\} .
\end{equation}

\begin{equation}
    L = d_{CD}(S, \bar{S_1}) + d_{CD}(S, \bar{S_2}) .
\end{equation}

\begin{equation}
    \hat{S} = D_2(D_1(E(S;\theta_e); \theta_{d1}); \theta_{d2}) .
\end{equation}

We use the trained encoder $E$ to extract the latent vector $z$, i.e., $z = E(S; \theta_e)$. This latent vector is a more discriminative representation than the form of a 3D point cloud, and we found it can be efficiently classified using one-class classifiers.

\subsubsection{Training for the general feature extractor.}
The point cloud autoencoder is trained on a dataset that is independent of normal and anomaly class distributions and is used as a general feature extractor. By training the autoencoder to reconstruct various shapes of the input data, it can extract shape features of unseen classes. We observed that both CAD models~\cite{ModelNet} and mathematically generated fractal 3D structures~\cite{Fractal} serve as effective encoder training datasets.

\begin{figure}
\begin{center}
\includegraphics[width=0.9\textwidth]{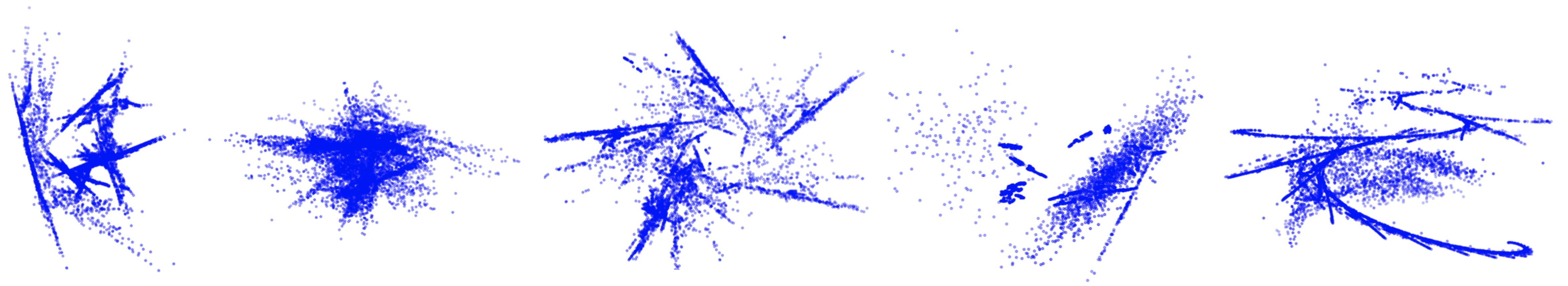}
\caption{Auto-generated 3D fractal point clouds based on the formula of \cite{Fractal}.}
\label{fig:fractal}
\end{center}
\vspace{-0.3cm}  
\end{figure}

\subsection{Latent-based One-class Classifiers}
The trained general feature extractor transforms the input 3D point cloud into a 512-dimensional latent vector $z$ at the bottleneck of the feature extractor where the local and global shape features condense. The extracted latent vectors of training data is used to train an unsupervised one-class classification approach. Note that, only normal data is available during training of the classifier. We compare several one-class classification techniques, including OC-SVM~\cite{OCSVM}, GODS~\cite{GODS}, DeepSVDD\cite{DeepSVDD}, KPCA-ND~\cite{KPCA-ND} in order to validate which classifier is appropriate for identifying the point cloud latent vectors. 

In OC-SVM, kernel OC-SVM with RBF kernel is used to enhance discrimination. The Gaussian kernel is used for KPCA-ND to map the features to an infinite-dimensional feature. For DeepSVDD, we replaced the convolution layers with several connected layers to handle 512-dimensional latent vectors as the network input. For GODS, we used the higher number of subspace hyperplanes to increase the discrimination power in the latent space where the shape features are nonlinearly expressed.

\section{Experiments}

\subsection{Training of the General Feature Extractor}
In our point cloud autoencoder, the input and output sizes of 3D point clouds are set to 2048. We set the learning rate to 0.0001, the training epochs to 300, and Chamfer distance as the reconstruction loss, following the work of~\cite{keio}. Once the autoencoder is trained, the weights are fixed and the frozen autoencoder is commonly used in testing all the datasets. 

\subsubsection{Encoder-training Datasets}
We compare the performance of the encoder training datasets - ModelNet10, ModelNet40~\cite{ModelNet}, Fractal400 and Fractal1000 as Fractal 3D point clouds~\cite{Fractal}. The fractal point clouds are generated by the 3D fractal models and mixed with fractal noise following~\cite{Fractal} with the variance threshold 0.05 and the noise ratio 0.2 (Figure~\ref{fig:fractal}). 
From each point cloud, 2048 points are randomly sampled. 

\begin{table}
\centering
\caption{Encoder-training datasets.}\label{tab1}
\begin{tabular}{l|l|l|l}
\hline
Encoder-training Dataset & Description & Number of classes & Sample size \\
\hline
ModelNet10 & 3D CAD object models & 10 & 3991\\
ModelNet40 & 3D CAD object models & 40 & 9843\\
Fractal400 & Fractal point clouds & 400 & 4000\\
Fractal1000 & Fractal point clouds & 1000 & 10000\\
\hline
\end{tabular}
\label{tbl:predatasets}
\vspace{-0.1cm}  
\end{table}

\subsection{Novelty Detection Datasets}
Following the work of~\cite{keio}, we perform novelty detection on the subsets of the point cloud dataset~\cite{ShapeNet} by defining one class as an anomaly and the rest classes of the subset as normal. The training data contains only a normal class, while test data includes both normal and anomaly classes. 
While the previous work~\cite{keio} uses only one subset of the ShapeNet dataset~\cite{ShapeNet} with seven classes, we prepare three different subsets to deeply evaluate the performances of each method. \\
\textbf{1) ShapeNet-7} is composed of seven classes in the ShapeNet dataset used in the previous work~\cite{keio}: Lamp, Chair, Table, Car, Sofa, Rifle, and Airplane. \\
\textbf{2) ShapeNet-small} includes seven classes with smaller sample sizes than the ShapeNet-7: Bookshelf, Laptop, Knife, Train, Motorbike, Guitar, and Faucet. \\
\textbf{3) ShapeNet-14} is the combined dataset of ShapNet-7 and ShapeNet-small with 14 classes: Lamp, Chair, Table, Car, Sofa, Rifle, Airplane, Bookshelf, Laptop, Knife, Train, Motorbike, Guitar, and Faucet. 

Each class contains hundreds of samples, and each sample has 2048 3D points that are randomly selected from the original point cloud. If the number of 3D points is smaller than 2048, they are randomly duplicated to reach 2048. The samples within each class are randomly divided into a training, validation, and test set at a ratio of 7:1:2. The training data is used to train the one-class classifier.

\subsection{Training and Inference of One-class Classifiers}
The extracted latent vectors of training data from the frozen autoencoder are utilized to train the one-class classifier. The test data is also transformed into latent vectors and then classified by the trained classifiers. OC-SVM and GODS compute the anomaly score by the normalized signed distance between the input latent vector and the learned hyperplane. KPCA-ND calculates the anomaly score by cosine similarity between the eigenvectors and the mapped latent vector. DeepSVDD defines the anomaly score directly using the anomaly class probabilities output by the classifier network.
The hyperparameters of each one-class classifier are determined through a grid search using the validation data, with a predefined range of each hyperparameter. A single set of selected hyperparameters is then used to evaluate each novelty detection.

\textbf{Evaluation} We evaluate each method by calculating AUC while varying the threshold value of the anomaly score according to the proposed methods ~\cite{eval1,eval2}.

\begin{table}[ht]
\centering
\caption{Parameters for the grid search.}
\vspace{-0.2cm}
\begin{tabular}{c|c}
\hline
Method & Parameter \\
\hline
{OC-SVM} & an upper bound on the fraction of training errors $\nu$, kernel coefficient $\gamma$ \\
\hline
\multirow{2}{*}{GODS} & the weight of controlling how far the predictions is from the hyperplanes  $\eta$, \\
                      & step size $\lambda$, the number of subspaces \\
\hline
\multirow{2}{*}{DeepSVDD} & an upper bound on the fraction of training data $\nu$, \\
                          & radius $r$, network, epoch \\
\hline
{KPCA-ND} & kernel width $\sigma$, the number of eigenvectors \\
\hline
\end{tabular}
\label{hyperparameters}
\vspace{-0.2cm}  
\end{table}

\begin{table*}[ht]
  \centering
  \caption{Comparison of novelty detection performance on ShapeNet-7. Each row represents the result where one class is defined as an anomaly class, and the rest classes are defined as normal. ``Baseline'' detects anomalies by the autoencoder reconstruction loss.}
    \begin{tabular}{c|c|c|cccc}
      \hline
      \hline
      \multirow{2}{*}{Anomaly class} & \multirow{2}{*}{Masuda et al.~\cite{keio}} & \multirow{2}{*}{Baseline} &  & \multicolumn{2}{c}{Latent-based Method (Ours)} &  \\
       &  &  & OC-SVM & KPCA-ND & DeepSVDD & GODS  \\
      \hline
      lamp     & 0.716 & 0.715 & \textbf{0.936} & 0.880 & 0.866 & 0.868 \\
      chair    & 0.689 & 0.643 & 0.789 & 0.721 & \textbf{0.831} & 0.658 \\
      table    & 0.787 & 0.723 & 0.774 & \textbf{0.920} & 0.898 & 0.712 \\
      car      & 0.409 & 0.399 & 0.798 & \textbf{0.923} & 0.841 & 0.362 \\
      sofa     & 0.716 & 0.764 & 0.717 & \textbf{0.769} & 0.765 & 0.702 \\
      rifle    & 0.587 & 0.689 & \textbf{0.941} & 0.916 & 0.915 & 0.720 \\
      airplane & 0.674 & 0.410 & 0.936 & \textbf{0.947} & 0.913 & 0.733 \\
      \hline

      average  & 0.654 & 0.620 & 0.841 & \textbf{0.868} & 0.861 & 0.679 \\
      \hline
      \hline
    \end{tabular}
    \vspace{0.1cm} 
    
    \label{table1}
    \vspace{-0.3cm}  
\end{table*}

\begin{table*}[ht]
  \centering
  \caption{Comparison of novelty detection performance on ShapeNet-small.}
    \begin{tabular}{c |c|c| c c c c }
      \hline
      \hline
      \multirow{2}{*}{Anomaly class} & \multirow{2}{*}{Masuda et al.~\cite{keio}} & \multirow{2}{*}{Baseline} &  & \multicolumn{2}{c}{Latent-based Method (Ours)} &  \\
       &  &  & OC-SVM & KPCA-ND & DeepSVDD & GODS  \\
      \hline
      bookshelf  & 0.886 & 0.903 & 0.911 & \textbf{0.926} & 0.857 & 0.855 \\
      laptop     & 0.930 & 0.904 & 0.964 & \textbf{0.982} & 0.956 & 0.984 \\
      knife      & 0.731 & 0.730 & \textbf{0.863} & 0.801 & 0.614 & 0.769 \\
      train      & 0.512 & 0.724 & \textbf{0.837} & 0.778 & 0.778 & 0.800 \\
      motorcycle & 0.373 & 0.450 & 0.890 & 0.932 & \textbf{0.941} & 0.826 \\
      guitar     & 0.561 & 0.735 & \textbf{0.919} & 0.874 & 0.885 & 0.810 \\
      faucet     & 0.881 & 0.843 & 0.954 & \textbf{0.982} & 0.927 & 0.950 \\
      \hline
      average    & 0.696 & 0.756 & \textbf{0.905} & 0.896 & 0.851 & 0.856 \\
      \hline
      \hline
    \end{tabular}
    \label{table2}
    \vspace{-0.1cm}  
\end{table*}

\begin{table*}[ht]
  \centering
  \caption{Comparison of novelty detection performance on ShapeNet-14 with Different classifiers.}
    \begin{tabular}{c | c | c | c c c c}
      \hline
      \hline
      \multirow{2}{*}{Anomaly class} & \multirow{2}{*}{Masuda et al.~\cite{keio}} & \multirow{2}{*}{Baseline} &  & \multicolumn{2}{c}{Latent-based Method (Ours)} &  \\
       &  &  & OC-SVM & KPCA-ND & DeepSVDD & GODS  \\
      \hline
      lamp       & 0.553 & 0.578 & \textbf{0.861} & 0.819 & 0.753 & 0.794 \\
      chair      & 0.605 & 0.636 & 0.660 & 0.747 & \textbf{0.773} & 0.526 \\
      table      & 0.616 & 0.750 & 0.703 & \textbf{0.868} & 0.858 & 0.593 \\
      car        & 0.309 & 0.527 & 0.783 & \textbf{0.931} & 0.848 & 0.677 \\
      sofa       & 0.771 & 0.723 & \textbf{0.818} & 0.760 & 0.793 & 0.702 \\
      rifle      & 0.521 & 0.636 & \textbf{0.929} & 0.836 & 0.855 & 0.729 \\
      airplane   & 0.421 & 0.621 & 0.907 & 0.914 & \textbf{0.920} & 0.799 \\
      bookshelf  & 0.566 & 0.566 & 0.855 & 0.860 & \textbf{0.904} & 0.463 \\
      laptop     & 0.771 & 0.903 & 0.939 & 0.943 & \textbf{0.955} & 0.927 \\
      knife      & 0.631 & 0.669 & \textbf{0.828} & 0.672 & 0.685 & 0.659 \\
      train      & 0.579 & 0.703 & \textbf{0.863} & 0.661 & 0.577 & 0.611 \\
      motorcycle & 0.357 & 0.501 & 0.900 & \textbf{0.951} & 0.891 & 0.455 \\
      guitar     & 0.583 & 0.555 & \textbf{0.904} & 0.840 & 0.892 & 0.804 \\
      faucet     & 0.786 & 0.695 & \textbf{0.914} & 0.841 & 0.800 & 0.858 \\
      \hline
      average    & 0.572 & 0.647 & \textbf{0.847} & 0.832 & 0.822 & 0.686 \\
      \hline
      \hline
    \end{tabular}
    \label{table3}
    \vspace{-0.3cm}  
\end{table*}

\subsection{Comparison}
We experiment with our latent-based approach with several one-class classifiers, where the autoencoder is trained on ModelNet10. Additionally, we compare our method against two different reconstruction-based methods, one using a variational autoencoder (Masuda et al.~\cite{keio}), and the other using a standard autoencoder that we set as the baseline. For \cite{keio} and baseline methods, we train the networks with the normal classes of each Shepenet subset, and use the chamfer distance as the anomaly score, since it showed better AUC than the other types of anomaly scores of~\cite{keio}. 
The measured AUC on Shapnent-7, ShapeNet-small, and ShapeNet-14 are shown in Table~\ref{table1}, Table~\ref{table2}, and Table~\ref{table3}. These results show that our latent-based methods outperform the reconstruction-based approaches across all the datasets, demonstrating the efficiency of one-class classification on latent vectors of 3D point clouds. Among the one-class classifiers, OC-SVM and KPCA-ND exhibit higher performance on average in all the datasets.

\begin{table}[ht]
  \centering
  \caption{Comparison of novelty detection performance on different encoder-training datasets.}
  \begin{tabular}{l|c|c|c|c|c|c|c|c}
   \hline
    \textbf{\scriptsize} & \multicolumn{8}{c}{Encoder-training Dataset}\\
    \hline
    \textbf{\scriptsize} & \multicolumn{2}{c|}{\textbf{ModelNet10}} & \multicolumn{2}{c|}{\textbf{ModelNet40}} & \multicolumn{2}{c|}{\textbf{Fractal400}} & \multicolumn{2}{c}{\textbf{Fractal1000}} \\
    \cline{2-9}
    & \scriptsize OC-SVM & \scriptsize KPCA-ND & \scriptsize OC-SVM & \scriptsize KPCA-ND & \scriptsize OC-SVM & \scriptsize KPCA-ND & \scriptsize OC-SVM & \scriptsize KPCA-ND \\
    \hline
    ShapeNet-7     & 0.841 & \textbf{0.868} & 0.813 & 0.851 & 0.784 & 0.846 & 0.820 & 0.865 \\
    ShapeNet-small & 0.905 & 0.896 & 0.885 & 0.888 & 0.881 & 0.876 & \textbf{0.915} & 0.908 \\
    ShapeNet-14    & 0.847 & 0.832 & 0.827 & 0.844 & 0.811 & 0.839 & 0.831 & \textbf{0.853} \\
    \hline
    average & 0.865 & 0.865 & 0.842 & 0.861 & 0.825 & 0.854 & 0.855 & \textbf{0.875} \\
    \hline
  \end{tabular}
  \label{table:combined}
\vspace{-0.2cm}  
\end{table}

\subsubsection{Comparison of Encoder-training Datasets}
We compare the performance of the general feature extractor with different encoder-training datasets, ModelNet10, ModelNet40, Fractal400, and Fractal1000. As shown in Table~\ref{tbl:predatasets}, Fractal1000 and ModelNet10 exhibited the highest AUC among the datasets.

\begin{figure}
\centering
\includegraphics[width=0.9\textwidth]{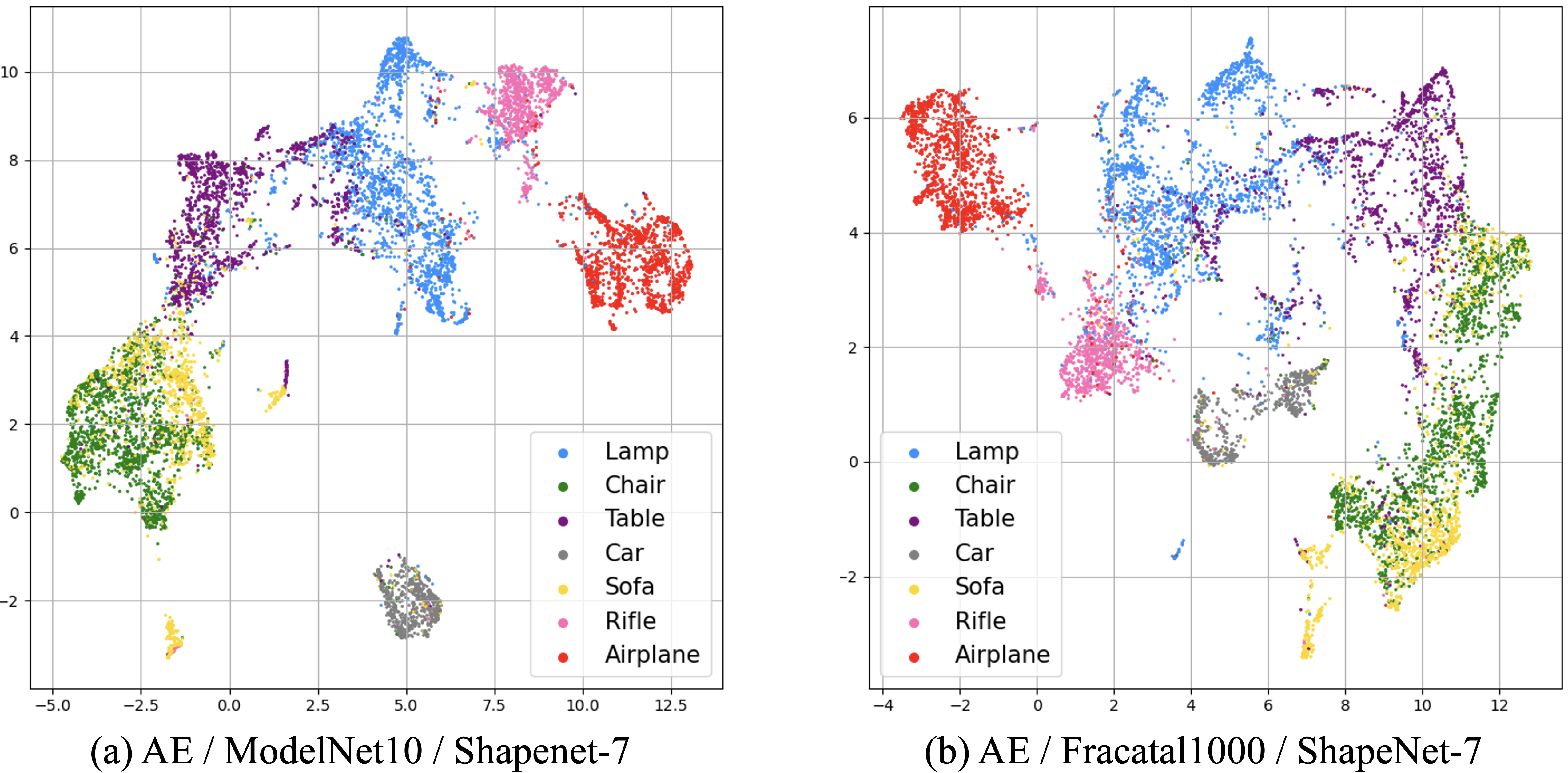}
    \caption{Latent vectors of ShapeNet-7 visualized using Umap~\cite{Umap}. (a) The latent vectors are extracted by the autoencoder trained on ModelNet10. 
    (b) The latent vectors are extracted by the autoencoder trained on Fractal1000.}
\label{fig:m10}
\vspace{-0.3cm}  
\end{figure}

\begin{figure}
\centering
\includegraphics[width=0.85\textwidth]{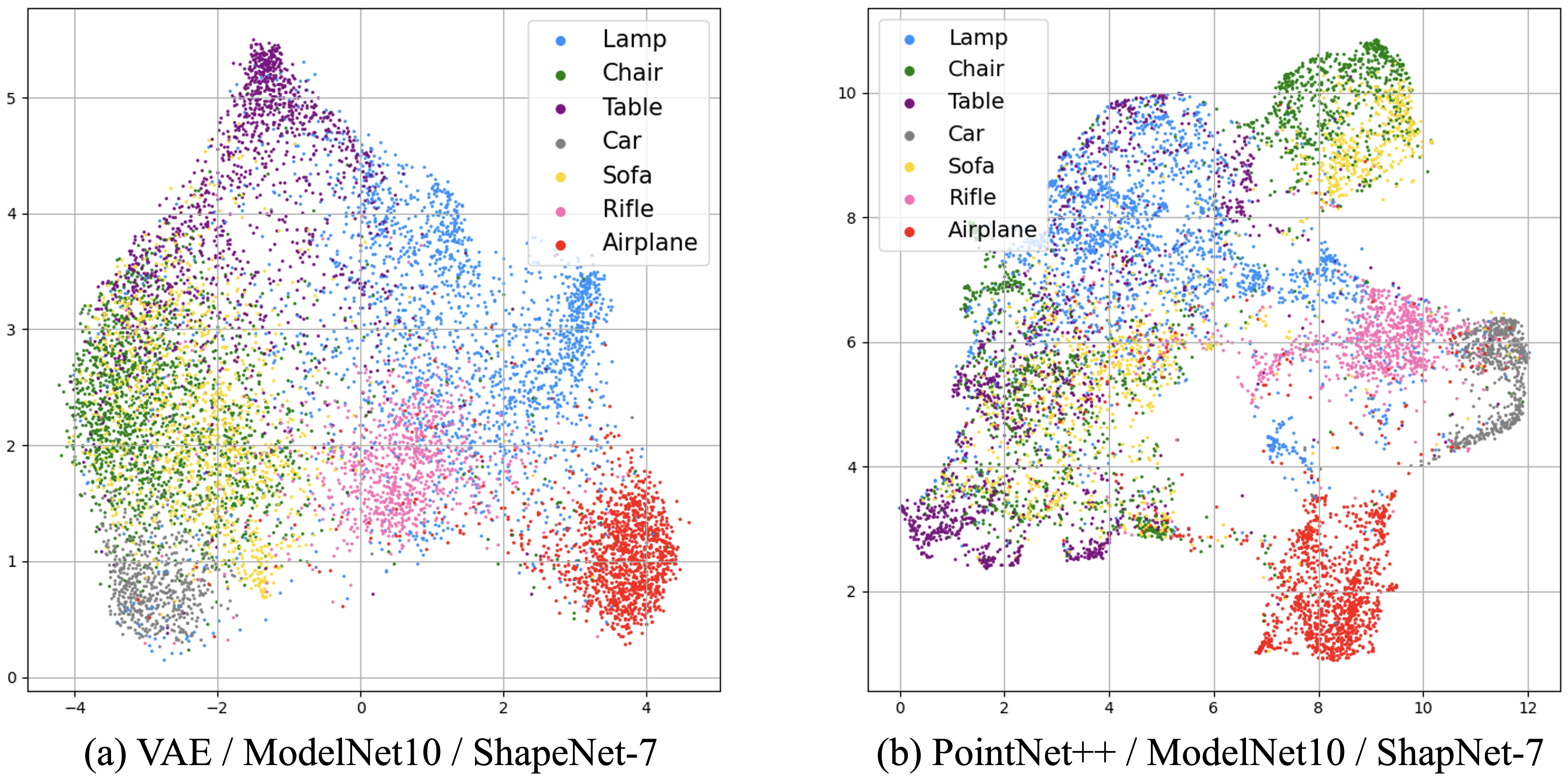}
\caption{Latent vectors of ShapeNet-7 visualized using Umap~\cite{Umap}. (a) The VAE~\cite{keio} trained on ModelNet10. (b) The PointNet++~\cite{PointNet2} trained on ModelNet10. Each color represents each class.}
\label{fig:VAE_PointNet}
\vspace{-0.5cm}  
\end{figure}

\subsection{Discussion}
The results demonstrate that our latent-based one-class classification using the general feature extractor outperforms existing reconstruction-based methods for point cloud novelty detection.

\textbf{Effects of general feature extractor.}
Figure~\ref{fig:m10} (a) illustrates the latent vectors of ShapeNet-7 that are extracted from the autoencoder trained on ModelNet10. The latent vectors show good separation within classes, suggesting the successful shape feature extraction in the latent space. This leads to effective one-class classification in the latent representations.

\textbf{Comparison of encoder-training datasets.} When comparing the encoder-training datasets, Fractal1000 and ModelNet10 exhibited greater AUC scores. While ModelNet10 has some similar shapes in ShapeNet, fractal geometric structures generated from formulas do not have any similar categories in ShapeNet, yet showed superior performance to the existing methods. Figure~\ref{fig:m10} (b) illustrates the latent vector distributions of ShapeNet-7. This shows that each class remains distinguishable even when utilizing Fractal geometric shapes for the encoder training, indicating the various shapes in fractal geometries allow for the training of the autoencoder to capture shape features from even unseen data. This characteristic can be beneficial to apply it to a wide range of novelty detection tasks with low computational burden. The second row of  Table~\ref{table:extractors1} shows the performance when the autoencoder is trained on the normal data of ShapeNet-7, that is, the feature extractor is trained each time by using the normal categories. The performance is measured by OC-SVM. The result demonstrates that the general feature extractor exhibits higher AUC, indicating the effectiveness of utilizing the general feature extractor. This result also highlights the advantage of our latent-based method over the reconstruction-based method when the same dataset is used between methods Table~\ref{table1}, Table~\ref{table:extractors1}.

\textbf{Importance of using AE.}
Table~\ref{table:extractors} presents the performance of different feature extractors: our autoencoder (AE), variational autoencoder (VAE) ~\cite{keio}, and PointNet++~\cite{PointNet2}. All the feature extractors are trained on ModelNet10, and the classification performance is measured by OC-SVM. The results demonstrate that our autoencoder outperforms the other feature extractors.
Figure~\ref{fig:VAE_PointNet} (a) illustrates the latent vectors extracted at the bottleneck of VAE,  (b) depicts the latent vectors extracted after the abstraction layers in PointNet++ trained for classification. Both figures reveal that the distributions of each class are closely intertwined, making the one-class classification task complicated. 
As PointNet++ is trained using supervised learning, the latent representations are learned in a way that emphasizes the separation of seen classes, which may not be suitable for extracting features of unseen classes if the feature is extracted in the middle layer. Also, VAE employs a probabilistic transformation, which can hinder clear latent vector separation among classes.

\textbf{Comparison of one-class classifiers.} The best classifier can vary depending on the dataset and the distribution of latent variables. However, the results show that OC-SVM and KPCA-ND performed better than other one-class classifiers. Both OC-SVM and KPCA-ND are simple but effective kernel-based approaches, which we assume are effective to classify novelties from normal samples in the non-linear latent space. While Deep SVDD showed also good performance, it is relatively sensitive to the training epoch, which may make it challenging to opt for parameters in practical applications. Further improvement of the network architecture for this task could improve the detection performance.

\textbf{Limitation of the proposal framework.}  Among the classes of the datasets, the anomaly classes ‘chair’ and `sofa' showed the lower performance. Figure 3 illustrates that the latent vector distribution of the ‘chair’ class is closer to that of the ‘sofa’ class compared to the other classes. This suggests that detection becomes challenging when there are minor shape differences and the latent vectors have similar features.
\begin{table}[ht]
\centering
\caption{Comparison of novelty detection performance on ShapeNet-7 with different feature extractors. The average AUC is measuerd.} 
\begin{tabular}{c|c|c|c}
\hline
Network & Dimension & Encoder-training Dataset & AUC \\
\hline
Our AE & 512×1 & ModelNet10 & \textbf{0.841} \\
VAE~\cite{keio} & 512×1 & ModelNet10 & 0.643 \\ 
PointNet++\cite{PointNet2} & 1024×1 & ModelNet10 & 0.749 \\
\hline
\end{tabular}
\label{table:extractors}
\vspace{-0.2cm}  
\end{table}

\begin{table}[ht]
\centering
\caption{Comparison of novelty detection performance on ShapeNet-7 with different encoder-training datasets. The average AUC is measuerd.}
\begin{tabular}{c|c|c}
\hline
Network & Encoder-training Dataset & AUC \\
\hline
Our AE & ModelNet10 & \textbf{0.841} \\
Our AE & Normal Classes of ShapeNet-7 & 0.817 \\
\hline
\end{tabular}
\label{table:extractors1}
\vspace{-0.3cm}  
\end{table}

\section{Conclusion}
This paper has proposed the point cloud novelty detection approach consisting of the general feature extractor and the one-class classifier.
The autoencoder trained on a synthetic dataset can be used as the general feature extractor agnostic to the normal and anomaly categories and converts the input data into the latent vectors that represent shape features. The one-class classifier is trained on the latent vectors and classifies inputs as normal or anomalous. We conduct experiments on the subsets of ShapeNet dataset and our approach achieves higher performance than the existing reconstruction-based novelty detection methods. Also, the experiments showed the effectiveness of using the general feature extractor and OC-SVM and KPCA-ND as the one-class classifier. In the future, we will expand our methods for 3D point cloud novelty detection in real scenes.

\vskip \baselineskip
\noindent
\textbf{Aknowlegement. }
This work is partly supported by the Japan Society for the Promotion of Science (JSPS) through KAKENHI Grant Number 20K14516. The computation of this work was in part carried out on the Multi-wavelength Data Analysis System operated by the Astronomy Data Center (ADC), National Astronomical Observatory of Japan.


\begin{thebibliography}{8}

\bibitem{novelty} Pimentel, M. A. F., Clifton, D. A., Clifton, L., Tarassenko, L.: A review of novelty detection. Signal Processing vol.99, 215--249 (2014). https://doi.org/10.1016/j.sigpro.2013.12.026
\bibitem{ND} Yang, J., Zhou, Z., Li, Y., Liu, Z.: Generalized Out-of-Distribution Detection: A Survey. Preprint at arXiv.2110.11334 (2021)
\bibitem{keio} Masuda, M., Hachiuma, R., Fujii, R., Saito, H., Sekikawa, Y.: Toward Unsupervised 3d Point Cloud Anomaly Detection Using Variational Autoencoder. In: 2021 IEEE International Conference on Image Processing (ICIP), pp.3118--3122, Anchorage, AK, USA (2021). doi: 10.1109/ICIP42928.2021.9506795.
\bibitem{VAE} An, J., Cho, S.: Variational autoencoder based anomaly detection using reconstruction probability. Special Lecture on IE (2015)

\bibitem{dpd} Urbach, D., Ben-Shabat, Y., Lindenbaum, M.: DPDist: Comparing Point Clouds Using Deep Point Cloud Distance. In: Computer Vision - ECCV 2020: 16th European Conference, pp.545--560, Glasgow, UK (2020). https://doi.org/10.1007/978-3-030-58621-8\_32

\bibitem{AE} Chen, Z., Yeo, C. K., Lee, B. S., Lau, C. T.: Autoencoderbased network anomaly detection. Wireless Telecommunications Symposium (WTS). 1--5 (2018). doi: 10.1109/WTS.2018.8363930

\bibitem{OCSVM} Sch\"{o}lkopf, B., Platt, J. C., Shawe-Taylor J. C., Smola, A. J., Williamson, R. C.: Estimating the support of a high-dimentional distribution. Neural Computation, 13(7), 1443--1471 (2001). https://doi.org/10.1162/089976601750264965

\bibitem{KPCA-ND} Hoffmann, H.: Kernel PCA for novelty detection. Pattern Recognition, 40(3), pp.863--874 (2007). https://doi.org/10.1016/j.patcog.2006.07.009

\bibitem{NDImage0} Sabokrou, M., Khalooei, M., Fathy, M., Adeli, E.: Adversarially Learned One-Class Classifier for Novelty Detection. In: 2018 IEEE/CVF Conference on Computer Vision and Pattern Recognition, pp.3379--3388, Salt Lake City, UT, USA (2018). doi: 10.1109/CVPR.2018.00356

\bibitem{NDImage1} Sastry, C. S., Oore, S.: Detecting out-of-distribution examples with gram matrices. In: Proceedings of the 37th International Conference on Machine Learning (ICML'20), 119, pp.8491--8501 (2020)

\bibitem{NDImage2} Tack, J., Mo, S., Jeong, J., Shin, J.: CSI: Novelty Detection via Contrastive Learning on Distributionally Shifted Instances. In: Advances in Neural Information Processing Systems, 33, pp.11839--11852 (2020)

\bibitem{NDImage3} Huang, R., Geng, A., Li, Y.: On the Importance of Gradients for Detecting Distributional Shifts in the Wild. In: Advances in Neural Information Processing Systems (2021)

\bibitem{NDImage4} Xuefeng, D., Wang, Z., Cai, M., Li, Y.: VOS: Learning What You Don’t Know by Virtual Outlier Synthesis. In: Proceedings of the International Conference on Learning Representations (2022)

\bibitem{NDImage5} Sun, Y., Guo, C., Li, Y.: ReAct: Out-of-distribution Detection With Rectified Activations. In: Advances in Neural Information Processing Systems (2021)

\bibitem{teacherstudent} Qin, J., Gu, C., Yu, J., Zhang, C.: Teacher–student network for 3D point cloud anomaly detection with few normal samples. Expert Systems with Applications, 228, 120371 (2023). https://doi.org/10.1016/j.eswa.2023.120371

\bibitem{ShapeNet} Chang, A. X., Funkhouser, T., Guibas, L., Hanrahan, P., Huang, Q., Li, Z., Savarese, S., Savva, M., Song, S., Su, H., Xiao, J., Yi, L., Yu, F.: ShapeNet: An Information-Rich 3D Model Repository. Preprint at arXiv.1512.03012 (2015)

\bibitem{PointNet} Charles, R., Su, H., Kaichun, M., Guibas, L.: PointNet: Deep Learning on Point Sets for 3D Classification and Segmentation. In: 2017 IEEE Conference on Computer Vision and Pattern Recognition (CVPR), pp.77--85, Honolulu, HI, USA (2017). doi: 10.1109/CVPR.2017.16

\bibitem{PointNet2} Qi, C. R., Yi, L., Su, H., Guibas, L. J.: PointNet++: deep hierarchical feature learning on point sets in a metric space. Preprint at arXiv:1706.02413 (2017)

\bibitem{so-net} Li, J., Chen, B. M., Lee, G. H.: SO-Net: Self-Organizing Network for Point Cloud Analysis. Preprint at arXiv:1803.04249 (2018)

\bibitem{pointcontrast} Xie, S., Gu, J., Guo, D., Qi, C.R., Guibas, L., Litany, O.: PointContrast: Unsupervised Pre-training for 3D Point Cloud Understanding. In: Vedaldi, A., Bischof, H., Brox, T., Frahm, JM. (eds) Computer Vision – ECCV 2020. Lecture Notes in Computer Science(), 12348, 574--591 (2020). https://doi.org/10.1007/978-3-030-58580-8\_34

\bibitem{SSL-1} Achituve, I., Maron, H., Chechik, G.: Self-Supervised Learning for Domain Adaptation on Point Clouds. In 2021 IEEE Winter Conference on Applications of Computer Vision (WACV), pp.123--133, Waikoloa, HI, USA (2021). doi: 10.1109/WACV48630.2021.00017

\bibitem{SPU-Net} Liu, X., Liu, X., Liu, Y. -S., Han, Z.: SPU-Net: Self-Supervised Point Cloud Upsampling by Coarse-to-Fine Reconstruction With Self-Projection Optimization. In: IEEE Transactions on Image Processing, vol.31, pp.4213--4226 (2022). doi: 10.1109/TIP.2022.3182266

\bibitem{foldingnet} Yang, Y., Feng, C., Shen, Y., Tian, D.: FoldingNet: Point Cloud Auto-Encoder via Deep Grid Deformation. In: 2018 IEEE/CVF Conference on Computer Vision and Pattern Recognition, pp.205--215, Salt Lake City, UT, USA (2018). doi: 10.1109/CVPR.2018.00029
\bibitem{MAP-VAE} Han, Z., Wang, X., Liu, Y-S., Zwicker, M.: Multi-Angle Point Cloud-VAE: Unsupervised Feature Learning for 3D Point Clouds from Multiple Angles by Joint Self-Reconstruction and Half-to-Half Prediction. Preprint at arXiv:1907.12704 (2019)

\bibitem{Fractal} Yamada, R., Kataoka, H., Chiba, N., Domae, Y., Ogata, T.: Point Cloud Pre-Training With Natural 3D Structures. In: 2022 IEEE/CVF Conference on Computer Vision and Pattern Recognition (CVPR), pp. 21251--21261, New Orleans, LA, USA (2022). doi: 10.1109/CVPR52688.2022.02060

\bibitem{nearest} Pokrajac, D., Lazarevic, A., Latecki, L.: Incremental local outlier detection for data streams. In: 2007 IEEE Symposium on Computational Intelligence and Data Mining, pp.504--515, Honolulu, HI, USA (2007). doi: 10.1109/CIDM.2007.368917

\bibitem{k-nn} Syed, Z., Saeed, M., Rubinfeld, I.: Identifying High-Risk Patients without Labeled Training Data: Anomaly Detection Methodologies to Predict Adverse Outcomes. In: AMIA Annual Symposium Proceedings, pp.772--776 (2010)

\bibitem{SVDD} Tax, D., Duin, R.: Support vector data description. Machine Learning vol.54, 45--66 (2004). https://doi.org/10.1016/S0167-8655(99)00087-2

\bibitem{DeepSVDD} Ruff, L., Vandermeulen, R., Goernitz, N., Deecke, L., Siddiqui, S. A., Binder, A., M\"{u}ller, E., Kloft, M.: Deep One-Class Classification. In: Dy, J., Krause, A. (eds.) Proceedings of the 35th International Conference on Machine Learning, vol.80, pp.4393--4402 (2018).

\bibitem{PCA} Jolliffe, I. T.: Principal Component Analysis. Springer Series in Statistics, 2nd edn. Springer New York, NY (2002). https://doi.org/10.1007/b98835

\bibitem{kernelpca} Sch\"{o}lkopf, B., Smola, A., M\"{u}ller, K.-R.: Nonlinear Component Analysis as a Kernel Eigenvalue Problem.  Neural Computation, 10(5), 1299–-1319 (1998). doi: 10.1162/089976698300017467

\bibitem{GODS} Wang, J., Cherian, A.: GODS: Generalized One-class Discriminative Subspaces for Anomaly Detection. Preprint at arXiv.1908.05884 (2019)

\bibitem{chamfer} Fan, H., Su, H., Guibas, L. J.: A Point Set Generation Network for 3D Object Reconstruction from a Single Image. In: 2017 IEEE Conference on Computer Vision and Pattern Recognition (CVPR), pp.2463--2471, Honolulu, HI, USA (2017). doi: 10.1109/CVPR.2017.264

\bibitem{eval1} Akcay, S., Atapour-Abarghouei, A., Breckon, T. P.: GANomaly: Semi-supervised Anomaly Detection via Adversarial Training. In: ACCV 2018, Springer International Publishing, 11363, pp.622--637 (2018) 

\bibitem{eval2} Kimura, D., Chaudhury, S., Narita, M., Munawar, A., Tachibana, R.: Adversarial Discriminative Attention for Robust Anomaly Detection. In: Proceedings of the 2020 IEEE/CVF Winter Conference on Applications of Computer Vision (WACV), pp.2172--2181 (2020). doi: 10.1109/WACV45572.2020.9093428

\bibitem{Umap} McInnes, L., Healy, J., Melville, J.: UMAP: Uniform Manifold Approximation and Projection for Dimension Reduction. Preprint at arXiv.1802.03426 (2018)

\bibitem{ModelNet} Wu, Z., Song, S., Khosla, A., Yu, F., Zhang, L., Tang, X., Xiao, J.:
3D ShapeNets: A Deep Representation for Volumetric Shapes. In: 2015 IEEE Conference on Computer Vision and Pattern Recognition (CVPR), pp.1912--1920, Boston, MA, USA (2015). doi: 10.1109/CVPR.2015.7298801

\bibitem{resnet} He, K., Zhang, X., Ren, S., Sun, J.: Deep Residual Learning for Image Recognition. In: 2016 IEEE Conference on Computer Vision and Pattern Recognition (CVPR), pp.770--778, Las Vegas, NV, USA (2016). doi: 10.1109/CVPR.2016.90

\bibitem{imagenet} Deng, J., Dong, W., Socher, R., Li, L. -J., Li, K., Fei-Fei, L.: ImageNet: A large-scale hierarchical image database. In: 2009 IEEE Conference on Computer Vision and Pattern Recognition, pp.248--255, Miami, FL, USA (2009). doi: 10.1109/CVPR.2009.5206848


\end{thebibliography}
\end{document}